\documentclass[a4paper,11pt, twocolumn]{article}
\usepackage{arxiv}
\usepackage{times}

\usepackage[utf8]{inputenc} 
\usepackage[T1]{fontenc}    
\usepackage{url}            
\usepackage{booktabs}       
\usepackage{amsfonts}       
\usepackage{microtype}      
\usepackage{graphicx}
\usepackage{moreverb}
\usepackage{natbib}
\usepackage[colorlinks,bookmarksopen,bookmarksnumbered,citecolor=red,urlcolor=red]{hyperref}

\newcommand\BibTeX{{\rmfamily B\kern-.05em \textsc{i\kern-.025em b}\kern-.08em
T\kern-.1667em\lower.7ex\hbox{E}\kern-.125emX}}

\usepackage{booktabs}
\usepackage{multirow}
\usepackage[inline]{enumitem}
\usepackage[most]{tcolorbox}
\usepackage{colortbl}

\renewcommand{\cite}{\citep} 

\begin{document}


\title{Seeing Candidates at Scale: Multimodal LLMs for Visual Political Communication on Instagram}

\author{
  Michael Achmann-Denkler \\
  Media Informatics Group \\
  University of Regensburg \\
  Regensburg, Germany \\
  \texttt{michael.achmann@ur.de} \\
  \And
  Mario Haim \\
  Department of Media and Communication \\
  Ludwig-Maximilians-Universität \\
  Munich, Germany \\
  \texttt{haim@ifkw.lmu.de} \\
  \And
  Christian Wolff \\
  Media Informatics Group \\
  University of Regensburg \\
  Regensburg, Germany \\
  \texttt{christian.wolff@ur.de} \\
}







\twocolumn[
  \begin{@twocolumnfalse}
\maketitle
\begin{abstract}
This study presents a computational case study that evaluates the capabilities of specialized machine learning models and emerging multimodal large language models for Visual Political Communication (VPC) analysis. Focusing on concentrated visibility in Instagram stories and posts during the 2021 German federal election campaign, we compare the performance of traditional computer vision models (\texttt{FaceNet512}, \texttt{RetinaFace}, \texttt{Google Cloud Vision}) with a multimodal large language model (\texttt{GPT-4o}) in identifying front-runner politicians and counting individuals in images. \texttt{GPT-4o} outperformed the other models, achieving a macro $F_1$-score of 0.89 for face recognition and 0.86 for person counting in stories. These findings demonstrate the potential of advanced AI systems to scale and refine visual content analysis in political communication while highlighting methodological considerations for future research.
\end{abstract}

\begin{center}
\small\itshape
An earlier version was presented at \#SMSociety 2024 (London). Evaluation uses \texttt{gpt-4o-2024-05-13}, state-of-the-art on MMMU at the time of the 2024 experiments.
\end{center}

\keywords{Visual Political Communication, Computational Social Science, Concentrated Visibility, Multimodal LLMs, Political Campaigns}
   \vspace{0.5cm}
  \end{@twocolumnfalse}
]

\section{Introduction}
Visual political communication (VPC) is increasingly central to modern election campaigns, particularly on platforms like Instagram \cite{Farkas2023-tk}. Research suggests that visuals, being processed faster and evoking stronger emotional responses, influence voter behavior more effectively than text \cite{Grabe2009-pe, Joo2022-bj, Geise2015-sl}. However, while traditional Instagram posts have been extensively studied, ephemeral content like stories, special posts that vanish after 24 hours, remains underexplored \cite{Bast2021-fh,Towner2022-oe}. Stories, with their emphasis on spontaneity and immediacy, offer new insights into how politicians engage with voters. 

Yet, analyzing the vast amount of visual content on social media presents methodological challenges. Traditional manual coding is time-consuming and limits the scale of analysis \cite{Joo2022-bj}. Recent advancements in computer vision and machine learning offer new possibilities for automated visual content analysis, but their application in political communication research is still emerging \cite{Peng2024-uq, Schwemmer2023-ti}. 

This study contributes to this emerging area by evaluating computational approaches for automating the analysis of visual political content. We compare the performance of pre-trained computer vision models (\texttt{RetinaFace}, \texttt{FaceNet512}, \texttt{Google Cloud Vision}) and a multimodal large language model (\texttt{GPT-4o}) in identifying politicians and counting individuals in Instagram content.

We use the 2021 German federal election as a case study to apply and assess these methods. The campaign marked a pivotal moment as Germany transitioned from Angela Merkel’s leadership and leaned heavily on digital platforms due to the COVID-19 pandemic \cite{righetti2022politische}. As a case study application, we focus on the concept of concentrated visibility—the strategic emphasis on individuals over parties in visual content \cite{Van_Aelst2012-me}-- and explore how front-runner candidates were visually featured in posts and stories across account types.

This paper’s central question is: How effectively can computational methods automate the analysis of visual political content on Instagram, specifically in identifying and measuring concentrated visibility? In addressing this, we aim to offer a replicable methodological contribution to computational VPC research, while also shedding light on platform-specific visual strategies during the 2021 election.

\subsection{Visual Political Communication}

Visual political communication (VPC) has received increasing scholarly attention over the past decade, driven by the rise of visual-first platforms such as Instagram \cite{Farkas2023-tk}. Scholars have emphasized that visuals are not merely supplementary to text but convey political messages in distinct ways: they are processed holistically, evoke stronger emotional responses, and can enhance memory retention depending on context \cite{Geise2015-sl, Grabe2009-pe}. Facial expressions and visual portrayals of candidates, for instance, can shape perceptions of competence and trustworthiness, even influencing vote decisions \cite{Joo2019-pj, Olivola2010-eo}.

This growing relevance has sparked renewed interest in visual strategies within political campaigns. In response to earlier calls to move beyond text-based analyses \cite{Schill2012-oa}, recent works such as \textit{Visual Political Communication} \cite{Veneti2019-ea} and the \textit{Research Handbook on Visual Politics} \cite{Veneti2023-ck} have helped establish VPC as a central concern in political communication research. The shift toward the visual is particularly evident on platforms like Instagram, where campaign dynamics are increasingly shaped by the logic of visual content \cite{Farkas2023-tk}.

In this context, the visual prominence of individual politicians becomes a crucial aspect of campaign strategy. Scholars describe this shift as part of a broader trend of political personalization, which refers to the increasing focus on individuals over parties or institutions \cite{Adam2010-zn}. This trend manifests in different forms: professional personalization highlights politicians’ competencies and achievements, while intimate personalization emphasizes their private lives or emotional authenticity \cite{Hermans2013-li}.

Personalization is also closely linked to the concept of presidentialization, which captures the growing emphasis on front-runner candidates as central figures in political narratives \cite{Poguntke2005-if, Simunjak2023-ms}. \citet{Van_Aelst2012-me} provide a useful framework here, distinguishing between individualization—the increased visibility of politicians as individuals—and privatization—the portrayal of their personal traits or private lives. These conceptual distinctions form the basis for identifying specific manifestations of personalization in visual communication, one of which is the construct of concentrated visibility.

\subsection{Concentrated Visibility}
\begin{figure}[ht]
    \centering
    \includegraphics[width=1\linewidth]{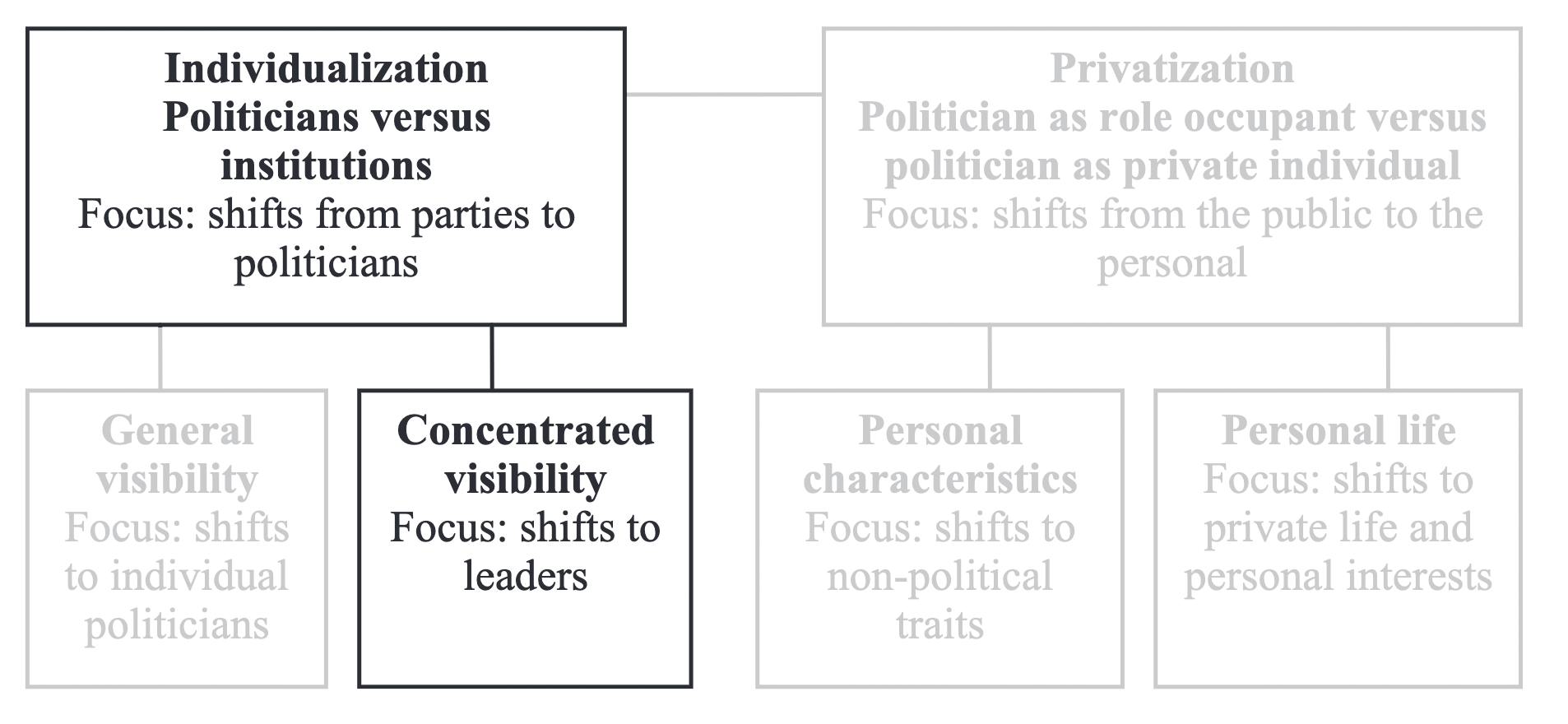}
    \caption{Dimensions of personalization according to \citet{Van_Aelst2012-me}. We highlighted the strand of `concentrated visibility', this study's focus.}
    \label{fig:concentrated-visibility}
\end{figure}

A central analytical lens in this study is concentrated visibility, a sub-dimension of individualization introduced by \citet{Van_Aelst2012-me} (cf. figure \ref{fig:concentrated-visibility}). It refers to the disproportionate focus on a select group of individuals—typically party leaders or front-runners—within campaign communication. This concept helps explain how visual narratives may become centered around key figures, reinforcing hierarchical structures of attention and political salience.

Concentrated visibility is particularly useful in visual media contexts, where repetition, prominence, and framing of individuals can convey symbolic meaning. These visual strategies often intersect with concepts of self-presentation and visual framing. For example, frequent appearances of a front-runner in formal settings can construct an “ideal candidate” image that signals authority and competence. In contrast, visuals featuring group interactions or casual, everyday moments may contribute to a populist framing, emphasizing accessibility and closeness to citizens \cite{Grabe2009-pe, Steffan2023-pd}.

These dynamics were evident in the 2021 German federal election. \citet{Steffan2023-pd}, building on \citeauthor{Grabe2009-pe}’s framework, identified different visual strategies among the top candidates: Armin Laschet primarily followed the “ideal candidate” frame; Annalena Baerbock leaned into a more populist visual self-personalization; and Olaf Scholz, who ultimately won the election, incorporated elements of both.

To capture such patterns systematically, we adopt a computational approach to measuring concentrated visibility. Using facial recognition and automated person counting, we quantify how often specific individuals appear in campaign content. This method enables large-scale, reproducible analyses of individualization and lays the groundwork for automated studies of self-presentation and visual framing.

\subsection{(Visual) Political Communication on Instagram}

While earlier studies of visual political communication primarily examined how traditional media portrayed candidates and parties, the rise of visual social media platforms has expanded the focus of research. Instagram's role in political communication has been extensively studied, addressing various political actors and nations. Studies commonly reveal that political figures use Instagram to project positive imagery rather than for policy discussion or voter engagement \cite{Bast2021-fh}. For instance, \citet{Bast2021-bi} examined the visual communication strategies of right-wing populist politicians on Instagram, identifying three key messaging strategies: professional statesmanship, personal self-presentation, and issue-focused posts. She found that personal self-presentation, which highlights the privatization of the political figure through informal and relatable imagery, resonates more with followers and fosters greater engagement.

Similarly, \citet{Liebhart2017-wf} analyzed the Instagram profile of Alexander van der Bellen's presidential campaign in Austria, identifying visual strategies that emphasized the candidate's biography, campaign team, and presentation as an eligible office holder. \citet{Ekman2017-kx} found that Swedish politicians use Instagram for visual self-communication, centering on the visual branding of individual politicians rather than interaction with the public. They argue that Instagram's logic revolves around the visual performativity of the 'self' \cite{Ekman2017-so}.

In the context of Germany, an analysis of visual political communication during the 2017 federal election showed that most posts shared by top candidates and parties featured individuals \cite{Hasler2021-gg}. Candidate accounts were identified as the primary location for presidentialization, with approximately one-third of party posts featuring the front-runners. However, \citeauthor{Hasler2021-gg} did not find indications of individualization or privatization strategies through Instagram posts during the hot phase of the campaign.

In their longitudinal study covering three German election campaigns, \citet{Hasler2023-kr} found that while most Instagram posts featured people, only about one-third focused on individual politicians alone. They argue that German political campaigns are not heavily centered on individual candidates. Furthermore, they dismiss presidentialization as the primary strategy, noting that half of the images did not feature a top candidate.

Regarding the 2021 German federal election campaign, \citet{Geise2024-un} found that politicians effectively leveraged personalization and emotionalization in their Facebook and Instagram campaigns. They applied multimodal strategies that integrated both images and texts. Their research highlights that Instagram was favored for more private and informal portrayals, while Facebook maintained a more formal tone. 

This aligns with \citet{Farkas2021-so}, who compared the visual communication strategies of Hungarian politicians between Facebook and Instagram, finding that content on both platforms was highly personalized, with Instagram content showing more personalization than Facebook. They observed that images featuring candidates were likelier to receive likes from users, indicating that personalization is a fundamental aspect of visual political communication.

Overall, the impact of personalization on user engagement has been shown across several studies: \citet{Bene2017-gr} analyzed Facebook posts shared by Hungarian politicians during the 2014 general election, concluding that personalization positively affects user engagement, resulting in more likes and comments. Similarly, \citet{Larsson2019-ic} examined whether personalization in visual communication leads to more successful posts, finding that personalization was an effective strategy for Finnish political accounts. In contrast, content and policy-focused posts garnered more likes in party accounts.

Gender is another factor of personalization that affects engagement: \citet{Brands2021-tz} found that female politicians on Instagram receive more engagement when projecting authority through formal attire and traditionally masculine traits, even though both male and female politicians tend to reference feminine issues.

These studies illustrate that while personalization is a common strategy on Instagram, its manifestation varies across national contexts and platforms. The visual focus on individual politicians is important in voter engagement and shaping political personas. However, despite the prevalent use of personalization, a gap remains in understanding how these strategies are employed across stories and whether complementary strategies between candidate and party accounts persist in this format. 
\subsection{Instagram Stories in Political Campaigns}

While most studies focus on permanent posts, Instagram Stories remain underexplored despite their widespread use and unique affordances. Building on this foundation, our case study focuses on concentrated visibility in Instagram stories and posts. Specifically, how front-runners were visually prioritized during the 2021 German federal election. As social media platforms evolve, so do the practices and expectations surrounding political communication: As \citet{Raynauld2023-qf} argue, politicians must strategically develop a digital habitus to adapt to the communication styles required by platforms like Instagram, and new formats like ephemeral stories. They emphasize that social media platforms shape campaign practices, necessitating politicians to adjust their strategies to align with platform-specific norms and audience expectations, or platform vernacular \cite{Gibbs2015-se}.

Notably, while most research on visual political communication (VPC) and Instagram has focused on posts, few studies have explored the use of ephemeral Instagram stories \cite{Bast2021-bi, Towner2024-gr, Towner2022-oe}, leaving a gap in our understanding of this format's role in political campaigns. They have been regarded as one of the main reasons for Instagram's success and were initially invented by Snapchat \cite{Leaver2020-cj}. Instagram stories are designed for mobile devices; users can upload images or videos and add text and other stickers to enhance their stories. Introduced in 2016, this format provides an opportunity for spontaneity and authenticity \cite{Towner2022-oe}, which may influence how politicians present themselves and interact with voters. 

In one of the few studies regarding stories, \citet{Towner2024-gr} analyzed 730 stories from 20 candidates and highlighted the limited use of Instagram's affordances, with campaigns focusing on professional and formal imagery, such as rallies and canvassing, rather than more personal or ``backstage'' content. Stories were geared toward mobilization messages and rarely used interactive features. Similarly, \citet{Towner2022-oe} analyzed the use of Instagram stories during the 2020 U.S. presidential campaigns, examining how candidates utilized this format to engage with voters. Their findings highlighted a strong focus on professional and politically centered visuals, with candidates predominantly featuring themselves in campaign rallies and speeches. Opportunities for visual privatization, such as ``behind-the-scenes'' content or informal glimpses into candidates’ lives, were notably underutilized. 

Addressing the gap in the German context, \citet{Achmann2023-tm} used topic modeling on Instagram stories, which indicated that stories tended to document rallies and the campaign trail rather than policy issues. Additionally, \citet{achmann-denkler-etal-2024-detecting} investigated mobilization strategies in Instagram stories and posts during the 2021 German election campaign, discovering significant differences in how calls to action were used, with stories containing fewer calls to action than posts. Building on the same corpus, \citet{Achmann-Denkler2025-ne} applied a hierarchical decision tree and prompted GPT-4o to classify visual image types in stories and posts, finding that campaign events and collages dominate the visual layer of the campaign.

These findings from the previous election campaign give a first impression of how Instagram stories are used in election campaigns. \citet{Parmelee2023-ft} conducted an interview study with Generation Z young adults, who expressed a desire for politicians to use Instagram stories for personalized backstage content and interaction, such as answering questions from story reactions \cite{Parmelee2023-ft}. Despite limitations like a small sample size and U.S. focus, the study suggests that young voters want more interactive and personalized political content on social media.

Content analyses of private Instagram stories further support this notion. \citeauthor{Bainotti2020-rn} collected 292 stories from 15 users, finding that a third were self-representational and 20\% were selfies \cite{Bainotti2020-rn}. They identified two grammars: documentation (sharing events or moments) and interaction (creating engagement). \citeauthor{Georgakopoulou2021-pc} also notes that Instagram and Snapchat stories emphasize visuality and everyday sharing, aligning with these grammars \cite{Georgakopoulou2021-pc}.

The similarities between private Instagram story practices and young voters' desires suggest that platform practices shape user expectations. This implies that campaign stories may fulfill these expectations by focusing on front-runners and encouraging interaction.

Our study addresses this gap by examining stories and posts. Leveraging computational methods, we provide insights into the strategic use of concentrated visibility across these formats and how party and front-runner accounts complement each other. In doing so, we contribute to the broader understanding of how social media platforms shape election campaigns and how political communication strategies align with platform-specific practices and user expectations.

\subsection{The Role of Computational Methods in Visual Analysis}
In this section, we situate our approach within the broader methodological landscape of computational visual analysis, highlighting its relevance and limitations in studying political communication on social media. The application of computational approaches to analyzing (social media) images is still in its early stages \cite{Peng2024-uq}. These computational efforts, often described as Automated Visual Content Analysis \cite{Schwemmer2023-ti, Peng2024-uq, Araujo2020-kd}, Images as Data \cite{Joo2022-bj, Peng2024-uq}, or Distant Viewing \cite{Arnold2023-ev}, aim to bridge the gap between manual image coding and scalable, algorithm-driven analysis. According to \citeauthor{Peng2024-uq}, automated visual analysis "involves the use of computer algorithms and tools to replicate human abilities to extract meaningful information and insights from visual media." \cite{Peng2024-uq}. Advances in computer vision and machine learning have made it increasingly feasible to extract and classify objects and concepts from visual data. Previously, the manual coding of images by humans was both time-consuming and expensive \cite{Joo2022-bj}. Computational coding offers the capability to analyze much larger corpora, potentially uncovering patterns hidden in smaller datasets \cite{Peng2024-uq}. As \citeauthor{Manovich2020-qq} envisions, this allows researchers ``to see one billion images'' \cite[][p.~1]{Manovich2020-qq} -- to grasp visual culture at scale through computational means.

Despite these advancements, many social science studies of social media images still rely on relatively small, manually collected datasets, with few adopting machine learning or computer vision techniques at scale \cite{Chen2023-pj}. The growing volume of visual content on platforms such as Instagram and TikTok highlights the increasing importance of computational methods for analyzing visual data \cite{Peng2023-hm}.

This notion aligns with the concept of distant viewing, which emphasizes the role of computer vision as a method to study visual datasets at scale. However, \citeauthor{Arnold2023-ev} caution against the false promise of objectivity in computational approaches, noting that human biases, shaped by beliefs, values, and cultural ideologies, are inherently present in both the algorithms and the annotated training data used to develop these models \cite[][ch.~1]{Arnold2023-ev}. For example, commercial computer vision products have exhibited gender recognition biases \cite{Schwemmer2020-qn}. Controlling for biases and validating automated codings through human annotations thus remain critical steps in computational visual analysis \cite{Peng2024-uq, Araujo2020-kd}.

Another challenge involves the ``sensory gap'' \cite{Smeulders2000-fo} between theoretical concepts and automated analysis. While current computer vision tools offer applications such as object detection and face recognition, mapping the raw outputs of these models to theoretical constructs continues to challenge researchers in the field \cite{Peng2024-uq}.

Common computational tasks in the analysis of visual political communication include face and object detection, face recognition, and image captioning \cite{Peng2024-uq}. To illustrate the application of these tools in political communication research, we outline key studies employing both commercial APIs and custom computational approaches. Several studies have utilized commercial APIs for automated image analysis. For instance, \citeauthor{Joo2022-bj} leveraged the \texttt{Google Cloud Vision} API to examine differences in visual self-presentation among candidates during the 2018 U.S. general election \cite{Joo2022-bj}. Similarly, \citet{Haim2021-bp} employed \texttt{Face++}, a commercial face detection and recognition software, to analyze over 13,000 politicians' self-depictions and news portrayals, providing insights into media representation \cite{Haim2021-bp}.

\citet{Bossetta2023-mc} conducted a cross-platform analysis of political campaign images shared on Facebook and Instagram during the 2020 U.S. election. They employed \texttt{Amazon Rekognition}'s facial recognition and emotion detection API to classify emotions in politicians' facial expressions.  To ensure accuracy, they validated their computational approach by comparing API results with crowd-sourced annotations, highlighting the importance of validation when using automated tools.

Beyond commercial APIs, several studies have demonstrated the utility of custom computational approaches in analyzing visual political communication. For example, \citet{Peng2021-xy} used computer vision techniques to detect and identify faces in Instagram posts by U.S. politicians, finding that images featuring politicians were significant drivers of audience engagement. Similarly, \citet{Xi2020-sn} employed a convolutional neural network to classify legislators' political affiliations based on Facebook images, revealing how visual cues can convey political ideology. In the domain of video analysis, \citet{Tarr2022-ef} explored computational methods to automate the analysis of political campaign videos, including face detection and recognition techniques, achieving high accuracy in identifying candidate appearances. Likewise, \citet{Juergens2022-oh} developed a custom deep learning pipeline to detect faces, estimate age and gender, and analyze over 5,600 hours of German television, revealing structural inequalities in gender and age representation across broadcasters and genres. Their work illustrates the scalability and analytical depth of tailored computer vision methods beyond social media contexts.  Finally, \citet{Girbau2024-st} used the \texttt{FaceNet} model to analyze political figures in news archives, enabling flexible and efficient face detection without retraining.

Building on these approaches, we identified three types of models to computationally measure concentrated visibility: the open-source combination of \texttt{RetinaFace} and \texttt{FaceNet512}, the \texttt{Google Cloud Vision} API, and the multimodal Large Language Model \texttt{GPT-4o}.

\texttt{GPT-4o} is a multimodal Large Language Model (LLM).\footnote{\href{https://openai.com/index/hello-gpt-4o/}{https://openai.com/index/hello-gpt-4o/ (Accessed 2024-10-23).}} These relatively new models gained popularity quickly due to their versatility. They have demonstrated capabilities in social science, performing many tasks zero-shot, meaning without training data \cite{Ziems2024-yw}. \citet{Peng2024-uq} highlight the potential of multimodal models in advancing automated visual analysis. We chose \texttt{GPT-4o} for our experiments because, at the time of evaluation, it was the top-performing model on the MMMU benchmark \cite{Yue2023-rj} leaderboard, a widely used benchmark for evaluating multimodal understanding. 

Our comparison aims to determine the optimal models for effectively identifying politicians and counting individuals in Instagram posts and stories, contributing to advancements in automated visual content analysis. By bridging existing gaps in computational visual analysis, we offer a scalable approach for studying concentrated visibility in political communication, paving the way for future studies to explore aspects such as visual frames and self-presentation.

\subsection{Research Questions}
Building on the theoretical and methodological gaps outlined in the previous sections, this study addresses two main objectives. First, we evaluate computational approaches for automating visual political communication analysis, focusing on their performance in identifying individual politicians and counting individuals in Instagram stories and posts. This responds to the growing need for scalable methods in visual content analysis, as highlighted by previous research \cite{Peng2024-uq, Schwemmer2023-ti}.

\begin{description} 
\item[RQ1a] How do \texttt{FaceNet512} and \texttt{GPT-4o} compare in terms of classification performance when identifying politicians in Instagram stories and posts? 
\item[RQ1b] How do \texttt{RetinaFace}, the \texttt{Google Cloud Vision} API, and \texttt{GPT-4o} compare to human annotations in terms of classification performance when counting individuals pictured in Instagram stories and posts? \end{description}

Second, we explore patterns of concentrated visibility as a case study to investigate how visual communication strategies vary across different account types and political parties. This inquiry addresses the lack of research on ephemeral formats like Instagram stories and their role in shaping perceptions of political actors \cite{Bast2021-fh, Towner2022-oe}.

\begin{description} 
\item[RQ2a] How do the patterns of concentrated visibility vary between party and front-runner accounts and across different political parties? 
\item[RQ2b] How do the patterns of concentrated visibility differ between Instagram stories and posts? \end{description}

By bridging these methodological and empirical questions, we aim to advance computational methods for visual political communication research and provide new insights into campaign strategies during the 2021 German federal election.

\section{Methods}
For our case study, we measured concentrated visibility across posts and stories shared during the 2021 campaign in Germany. We measured concentrated visibility as proposed in previous studies \cite{Hasler2023-kr,Hasler2021-gg,Bast2021-bi,Russmann2019-uw} by a) determining the presence of front-runners in images and the first frame of videos and b) detecting whether they are pictured by themselves, or with others. Concentrated visibility is a subtype of personalization as constructed by \citet{Van_Aelst2012-me} that refers to the media's focused attention on a limited number of political leaders, often at the expense of broader coverage of other politicians or parties.

\begin{figure}
    \centering
    \includegraphics[width=.5\linewidth]{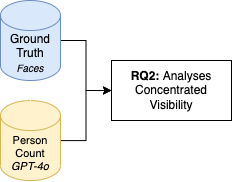}
    \caption{To measure concentrated visibility, we use the human annotations with the name labels for each face, and pair it with the person count based on the GPT-4o prompt.}
    \label{fig:workflow-rq2}
\end{figure}

This operationalization results in two classification tasks for the proposed automated coding approach: a) we need to check each image for the presence of front-runners, and b) we need to determine the number of people in the focus of each image. In this section, we report on our data collection process, describe each classification task in detail, and outline the evaluation of our computational approaches using human annotations for external validation. To answer RQ2a and RQ2b, questions about the patterns of concentrated visibility in the 2021 federal election campaign, we combine the dataset of human-annotated faces with the dataset of person counts produced by the best-performing model, as illustrated in figure \ref{fig:workflow-rq2}.

\subsection{Data Collection \& Corpus}
We collected 1,424 Instagram stories and 547 posts. The corpus comprises images and videos published by five parties (CDU, CSU, SPD, The Greens, and FDP) and one chancellor candidate/front-runner per party (see table \ref{tab:politicians}).\footnote{In the German political system, only major parties like CDU and SPD nominate a chancellor candidate, reflecting their potential to lead the government. Smaller parties, such as FDP, typically designate a front-runner ('Spitzenkandidat') to lead their campaign without aspiring to the chancellorship.} Data collection occurred from Sept. 12th until Sept. 25, 2021, excluding election day. Posts have been collected retrospectively using CrowdTangle, fetching additional images and videos using \texttt{instaloader}. Stories have been collected daily at 0:00 using the \texttt{selenium} Python package to simulate a human user browsing the stories.\footnote{We can not guarantee completeness for Sep 14 due to technical problems.} 

Since a large share of posts and stories were videos, we extracted the first frame of each video and treated it as a static image to keep the analysis manageable. Permanent posts may include multiple images or videos in a gallery post, while stories consist of precisely one image/video. Thus, our corpus consists of 957 images linked to 547 posts. Our unit of analysis is the image level unless stated otherwise. 

\begin{table}
\centering
\caption{Selected politicians' accounts and their positions and party affiliation at the time of data collection. (The party GRÜNE is referenced as B90DieGruenen later on.)}
\label{tab:politicians}
\resizebox{\linewidth}{!}{%
\begin{tabular}{@{}llll@{}}
\toprule
Name                    & Party     & Position             & @handle                   \\ \midrule
Armin Laschet           & CDU       & Chancellor Candidate & @armin\_laschet           \\
Annalena Baerbock       & GRÜNE     & Chancellor Candidate & @abaerbock                \\
Olaf Scholz             & SPD       & Chancellor Candidate & @olafscholz               \\
Christian Lindner       & FDP       & Front-Runner         & @christianlindner         \\
Markus Söder            & CSU       & Front-Runner         & @markus.soeder            \\ \bottomrule
\end{tabular} }%
\end{table}

\subsection{Face Recognition}

We compared two approaches to determine the presence of front-runners in images: We tested a workflow of face detection and face verification based on the deep learning models \texttt{RetinaFace} and \texttt{FaceNet512} against prompting using the multimodal large language model \texttt{GPT-4o}. 

\begin{figure}
    \centering
    \includegraphics[width=1\linewidth]{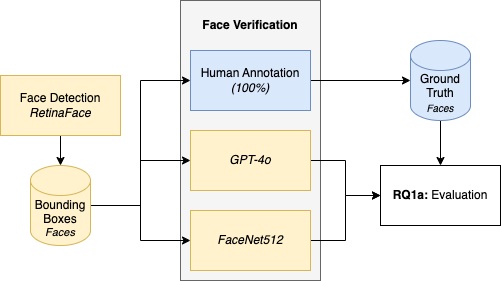}
    \caption{The face recognition workflow starts with face detection. The resulting bounding boxes were the basis for both the computational face search and the human annotations. In this study, human participants labeled all images (and bounding boxes). Comparing this ground truth to the computational results, we determined the classification quality of FaceNet512 and GPT-4o.}
    \label{fig:workflow-face-recognition}
\end{figure}

Determining the presence of the front-runners using the deep learning approach is a three-step process: First, the face detection model returns the coordinates for each face present in a given image. In the second step, the face recognition model compares every single face with one or multiple \textit{known} faces (see overview in figure \ref{fig:workflow-face-recognition}). This results in a label (=name) for each detected face. As we are interested in the presence of top candidates on the image level, we aggregate the individual face recognition results to determine the presence of any front-runners. In this final step, the labels assigned to the detected faces, either the front-runners' names or ``Unknown,'' are analyzed. 

We conducted our study with the help of the \texttt{deepface} \cite{serengil2020lightface} python library. This software offers a uniform interface for multiple open-source face detection and recognition models. Face detection and recognition were executed in \textit{Google Colab} notebooks utilizing \textit{Nvidia-T4} GPUs. 

We used the \texttt{RetinaFace} detection model \cite{Deng2020-fa} for the face recognition step and \texttt{FaceNet512} \cite{Schroff2015-tl} as our face recognition model. The \texttt{FaceNet512} model is a state-of-the-art deep learning face recognition model, which is reported to achieve one of the highest accuracies when tested using the \textit{Labeled Faces in the Wild} (LFW) dataset \cite{Huang2008-oh,Wang2021-ex}. The \texttt{deepface} authors report this combination as the best-performing one in their benchmark, surpassing human-level accuracy
\cite{serengil2024lightface}. The \texttt{RetinaFace} model is reported to overperform, which is favorable for preparing the annotation dataset while possibly confounding the person count.

One of the authors manually prepared a database of known faces for each front-runner to prepare the face search, with an average of 15 target images per candidate. The target images were removed from the dataset to maintain an independent corpus during evaluation. We picked frontal images with sufficient quality for identification and added one high-quality portrait per candidate from the parliament's website to the database.

We generated vector representations of all face images in the target database and the corpus using the \texttt{deepface} package and the \texttt{FaceNet512} model. To determine the label for each face, \texttt{deepface} compares these representations for input images with the database and returns the closest match. Using our ground truth data, we employed the Euclidean L2 distance measure and dynamically identified the optimal cut-off threshold for matches. The threshold was determined by maximizing the macro $F_1$ score, resulting in $0.950$ for our story dataset and $0.941$ for posts. Matches with a distance higher than the threshold were treated as ``Unknown'' faces. 

Preliminary experiments showed that the face recognition system's runtime efficiency and accuracy improved when we grouped images by party and compared each image to only the front-runner of the same party. This approach reduced false positive rates (wrongly identified Front-Runners) and maintained the objective of quantifying concentrated visibility per party.

\begin{figure*}[t]
    \begin{tcolorbox}[enhanced, drop shadow, colback=white, sharp corners=southwest, rounded corners=northeast, boxrule=1pt, width=\linewidth]
    \begin{minipage}{\linewidth}
    Imagine you are a computational social science PhD candidate. Your objective is to measure the concentrated visibility of political leaders in the 2021 Federal German election campaign on social media. Look at the image and determine if it depicts \texttt{NAME}, the front-runner of the \texttt{PARTY} party. Note: We are \textbf{not} interested in identifying individuals beyond this scope; our focus is solely on the presence or absence of the \texttt{PARTY}'s front-runner. \\
    
    First, provide a comment explaining whether the image clearly depicts the \texttt{PARTY} front-runner, does not depict the front-runner, or if you are unsure. Based on this, set the `pictures\_candidate' value as follows:
    \begin{enumerate}
        \item \textbf{True}: if the image clearly depicts the \texttt{PARTY} front-runner
        \item \textbf{False}: if the image does not depict the \texttt{PARTY} front-runner
        \item \textbf{Unsure}: if you cannot clearly determine if it is the \texttt{PARTY} front-runner
    \end{enumerate}
    
    Please return a JSON dictionary formatted as follows:
    \begin{verbatim}
    {
      "comment": "",
      "pictures_candidate": ["True", "False", "Unsure"]
    }
    \end{verbatim}
    \end{minipage}
    \end{tcolorbox}
    \caption{The prompt used to determine the presence of a political leader in social media images. The \texttt{NAME} and \texttt{PARTY} were replaced based on the post's/story's metadata. Having the model generate a comment for each request helped from the prompt optimization stage through the final analysis.}
    \label{fig:prompt-concentrated-visibility}
\end{figure*}

The second approach evaluated in our study, \texttt{GPT-4o}, worked differently. For the multimodal large language model, we created a text prompt requesting the model to verify the presence of a given front-runner. We started with a few hand-picked images and a simple hand-crafted prompt on the ChatGPT platform. We requested improved versions of our preliminary prompts on the chat platform through several iterations and settled for the prompt in Figure \ref{fig:prompt-concentrated-visibility}. \citet{Törnberg_2024}'s workflow for LLM usage suggests a similar approach, and language models have already been shown to be effective prompt engineers \cite{Pryzant2023-ch}. Our prompt expects two answers from the model: An answer about the front-runner's presence and a comment on the decision. Asking the model about the presence of a specific politician prevented triggering the model's safety mechanism, asking the model to name the person usually returns ``I cannot identify individuals in images.'' We sent a single request for each image using these parameters: \texttt{gpt-4o-2024-05-13}, \texttt{temperature=0}, \texttt{top\_p=1}, and \texttt{max\_tokens=600}.

\subsection{Counting People}
We compared several computational approaches to accurately measure the number of visible people in our dataset. We evaluated: 1) \texttt{RetinaFace} \cite{Deng2020-fa}, a deep learning-based face detection model that has already been used during the face recognition step, 2) \textit{Google Vision API}, which offers comprehensive object detection capabilities, and 3) \texttt{GPT-4o}, a multimodal large language model with capabilities for context-aware detection. We chose the models to compare the effectiveness of face detection, object detection, and prompting as technologies to aid social science research, thereby contrasting deep learning with language models. 

Following previous work \cite[e.g.,][]{Hasler2023-kr, Bast2021-bi}, we categorized the person count variable into four groups: 0, 1, 2, and 3 or more. For this step, the dataset was filtered to include only images showing front-runners. \texttt{RetinaFace} returns a set of bounding boxes for each input image; by counting these bounding boxes, we inferred the number of faces displayed in a given image. Similarly, the \textit{Google Vision API} returns bounding boxes for detected objects in images. By specifically counting the bounding boxes labeled as \textit{Person}, we inferred the number of people visible in each image. Face detection using \texttt{RetinaFace} was conducted in a \textit{Google Colab} notebook utilizing an \textit{Nvidia-T4} GPU. For \textit{Google Cloud Vision} and \texttt{GPT-4o}, we sent one request per image. We crafted a prompt similar to the method in the face detection section for the language model: Initially, we began with a basic question and iteratively refined the prompt using ChatGPT and a small annotated test sample. For the OpenAI API requests, we used the following parameters: \texttt{gpt-4o-2024-05-13}, \texttt{temperature=0}, \texttt{top\_p=1}, and \texttt{max\_tokens=600}.

We expected the face detection model to count too many people, as it also detects faces in the image's background. We anticipated that object detection would work better than face detection because it requires more than just a face to detect a person. Similarly, we expected the LLM to perform comparably to object detection since we could craft a more specific prompt describing our goal of counting the primary subjects of the image. 

\begin{figure}[t]
    \begin{tcolorbox}[enhanced, drop shadow, colback=white, sharp corners=southwest, rounded corners=northeast, boxrule=1pt, width=\linewidth]
    \begin{minipage}{\textwidth}
    Imagine you are a social science researcher, conducting an analysis of the 2021 German Federal election campaign. The image shows the front-runner: \texttt{NAME}. We're interested in measuring the individualization of candidates in the election campaign. \\
    
    Assess the image and provide the following information:

    \begin{enumerate}
        \item How many people are in the focus of the image? Select the right choice: `GPT Count': \texttt{["0", "1", "2", "3+"]}
        \item Is there a crowd of spectators visible? Select the right choice: `GPT Crowd': \texttt{["True", "False"]}
    \end{enumerate}
    
    Respond in valid JSON only.
    \end{minipage}
    \end{tcolorbox}
      \caption{The prompt used in the (3b) scenario to count people in images using the auxiliary crowd variable. Remove the \textit{italic part} for the (3a) prompt. The \texttt{NAME} was replaced with the front-runner identified by the human annotators.}
  \label{fig:prompt-topics}
\end{figure}

\begin{figure}
    \centering
    \includegraphics[width=1\linewidth]{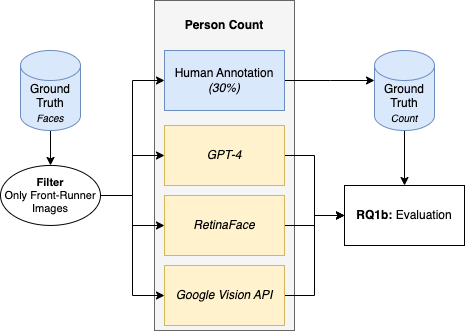}
    \caption{We filtered the corpus to include only images featuring front-runners based on human annotations. Human annotators marked 30\% of these images as a ground truth dataset for counting people. Using this dataset, we evaluated the results of three different models to determine the best approach.}
    \label{fig:workflow-3}
\end{figure}

\subsection{Annotation Studies}
We conducted an annotation study for each classification problem and image corpus: Human annotators coded the presence of front runners in stories and posts, and humans annotated the person count for every story and post image. 

To determine the presence of front-runners, 13 human annotators labeled the faces of politicians across all 1171 stories and 735 posts. Only images where RetinaFace identified faces were included in this study. Coders were recruited from our staff and students, the latter of whom were reimbursed through participant hours. The coders received an extensive manual with images for each sought-after politician. For stories, two annotators independently annotated each image using the Label Studio software. Following the high agreement for stories, post images were coded with a 30\% overlap. Annotators selected bounding boxes of front-runners and assigned predefined labels.

Disagreements between the coders were resolved through a review by one of the authors. We calculated Krippendorff's alpha \cite{Krippendorff1970-zr} to measure the interrater agreement to ensure the quality of our ground truth dataset. The agreement for both annotation studies was sufficient ($\alpha$ for stories: 0.86; for posts: 0.94). 

We conducted a separate annotation study to determine the person count. This time, we collected three independent annotations for each image. We used a filtered dataset in this study: Using the human annotations, we selected only images of front-runners.  Five annotators coded 30\% of these (193 stories, 145 posts). In line with the categorization of the computational count, the annotators coded the number of visible people (0, 1, 2, three, or more). We followed the same procedure as above to ensure consistency. Using the three annotations, we created the gold-standard dataset using the majority decision of all annotators. The agreement for both annotation studies was strong ($\alpha$ for stories: 0.81; for posts: 0.91). 

\subsection{Model Evaluations}
We used these annotations for the external validation of our computational approaches and evaluation of the models \cite{Birkenmaier2023-nt}. To determine the classification quality, we calculated the machine learning metrics accuracy, precision, recall, and --unless stated differently-- macro $F_1$ using \texttt{scikit-learn} \cite{scikit-learn}. We plotted confusion matrices for increased interpretability using \texttt{seaborn} \cite{Waskom2021}. For comparability with social science studies, we added the computational classifications as the n-th annotator to our annotation data and recalculated Krippendorff's $\alpha$ \cite{Haim2021-bp}. This way we can determine whether the model would increase or decrease the annotator agreement when considered an additional coder.

\section{Results: Model Evaluation}
\texttt{RetinaFace} detected a total of 4,738 faces in 1,171 stories (out of 1,424 images) and 3,533 faces in 708 images from 382 posts (out of 954 images and 547 posts). This indicates that 82.23\% of the collected stories and 69.84\% of the posts contain at least one face, with an average of 4.04 faces per story and 4.99 faces per post image. In the following section, we will first present the models' performance in face verification and person counting to address RQ1a and RQ1b. Subsequently, we will conduct a descriptive analysis of the personalization strategies used in posts and stories, followed by a statistical analysis of the observed patterns to answer RQ2a and RQ2b.

\subsection{RQ1a: Computational Face Recognition}

To address RQ1a, we compared the performance of two computational models—\texttt{FaceNet512} and \texttt{GPT-4o}—in recognizing political candidates' faces in Instagram stories and posts. Using human annotations as ground truth, we evaluated each model’s precision, recall, and $F_1$-scores to determine which provides more reliable results. For each party’s images, the model either correctly identified their front-runner’s presence (true positive, contributing to recall and precision), mistakenly labeled an unknown face as the candidate (false positive, affecting precision), or marked the candidate’s face as unknown (false negative, affecting recall). 

Table~\ref{tab:face-verification} summarizes the overall evaluation metrics for both models across stories and posts. The \texttt{GPT-4o} model outperformed \texttt{FaceNet512}, achieving higher macro and weighted $F_1$-scores, especially in stories. Specifically, \texttt{GPT-4o} attained a macro $F_1$-score of 0.89 for stories and 0.91 for posts, compared to \texttt{FaceNet512}’s 0.74 and 0.87, respectively.

\begin{table}[ht]

\caption{Comparison of the evaluation results for the face verification comparing FaceNet512 and GPT-4o. The table is split into \textbf{P}osts and \textbf{S}tories. We calculated Krippendorff's $\alpha$ to measure interrater agreement, incorporating the model as a fourth annotator to compare our results with social science research. }
\label{tab:face-verification}

\centering
\resizebox{\columnwidth}{!}{%
\begin{tabular}{l | c | c | c | c || c | c |}
\toprule
\textbf{Model} & \multicolumn{2}{c |}{\textbf{Macro $F_1$}} & \multicolumn{2}{c ||}{\textbf{Weighted $F_1$}}  & \multicolumn{2}{c |}{\textbf{Kripp. $\alpha$}} \\
                 & \textbf{S}     & \textbf{P}      & \textbf{S}       & \textbf{P}     &   \textbf{S}      & \textbf{P}       \\ \midrule
GPT-4o           & \textbf{0.89}  & \textbf{0.91}   &  \textbf{0.88}   & \textbf{0.91}  & 0.85              & 0.90              \\
FaceNet512      & 0.74           & 0.87            & 0.77             &  0.87          & 0.62              & 0.84              \\ \midrule
                                    \multicolumn{5}{r ||}{\textbf{Human Annotations}}    &  \textbf{0.87}    & \textbf{0.94}    \\
\bottomrule
\end{tabular}
}%
\end{table}

Examining the performance per candidate (see Table~\ref{tab:face-verification-detail} for stories), we observed that \texttt{GPT-4o} consistently provided higher precision and recall across all candidates in stories. For instance, Annalena Baerbock, the Green Party’s candidate, achieved an $F_1$-score of 0.91 with \texttt{GPT-4o}, significantly higher than the 0.66 obtained with \texttt{FaceNet512}. This improvement is particularly striking given that the specialized model achieved a recall of only 0.50 in detecting Baerbock.

\begin{table*}[ht]
\centering
\caption{Comparison of the \texttt{FaceNet512} and \texttt{GPT-4o} model verifying front-runners in stories. \textbf{Bold} values indicate the best-performing metric for each category and metric type.}
\label{tab:face-verification-detail}
\begin{tabular}{ l | c c c || c c c | c |}
\toprule
\multirow{2}{*}{\textbf{Person}} & \multicolumn{3}{c||}{\textbf{FaceNet512}} & \multicolumn{3}{c|}{\textbf{GPT-4o}} & \multirow{2}{*}{\textbf{Support}} \\
                & \textbf{Precision} & \textbf{Recall} & \textbf{F1-Score} & \textbf{Precision} & \textbf{Recall} & \textbf{F1-Score} &  \\
\midrule
Annalena Baerbock     & \textbf{0.98} & 0.50 & 0.66 & 0.91 & \textbf{0.90} & \textbf{0.91} & 118 \\
Armin Laschet         & \textbf{1.00} & 0.60 & 0.75 & 0.84 & \textbf{0.93} & \textbf{0.88} & 221 \\
Christian Lindner     & \textbf{0.97} & 0.54 & 0.69 & 0.88 & \textbf{0.92} & \textbf{0.90} & 61 \\
Markus Söder          & \textbf{0.98} & 0.60 & 0.75 & 0.89 & \textbf{0.87} & \textbf{0.88} & 68 \\
Olaf Scholz           & \textbf{0.95} & 0.63 & 0.76 & 0.85 & \textbf{0.90} & \textbf{0.88} & 117 \\
Unknown               & 0.69 & \textbf{0.99} & 0.81 & \textbf{0.90} & 0.85 & \textbf{0.88} & 555 \\
\midrule
\textbf{Macro Avg}    & \textbf{0.93} & 0.64 & 0.74 & 0.88 & \textbf{0.89} & \textbf{0.89} & 1140 \\
\textbf{Weighted Avg} & 0.84 & 0.78 & 0.77 & 0.88 & \textbf{0.88} & \textbf{0.88} & 1140 \\
\bottomrule
\end{tabular}
\end{table*}

\begin{figure*}
    \centering
    \includegraphics[width=1\linewidth]{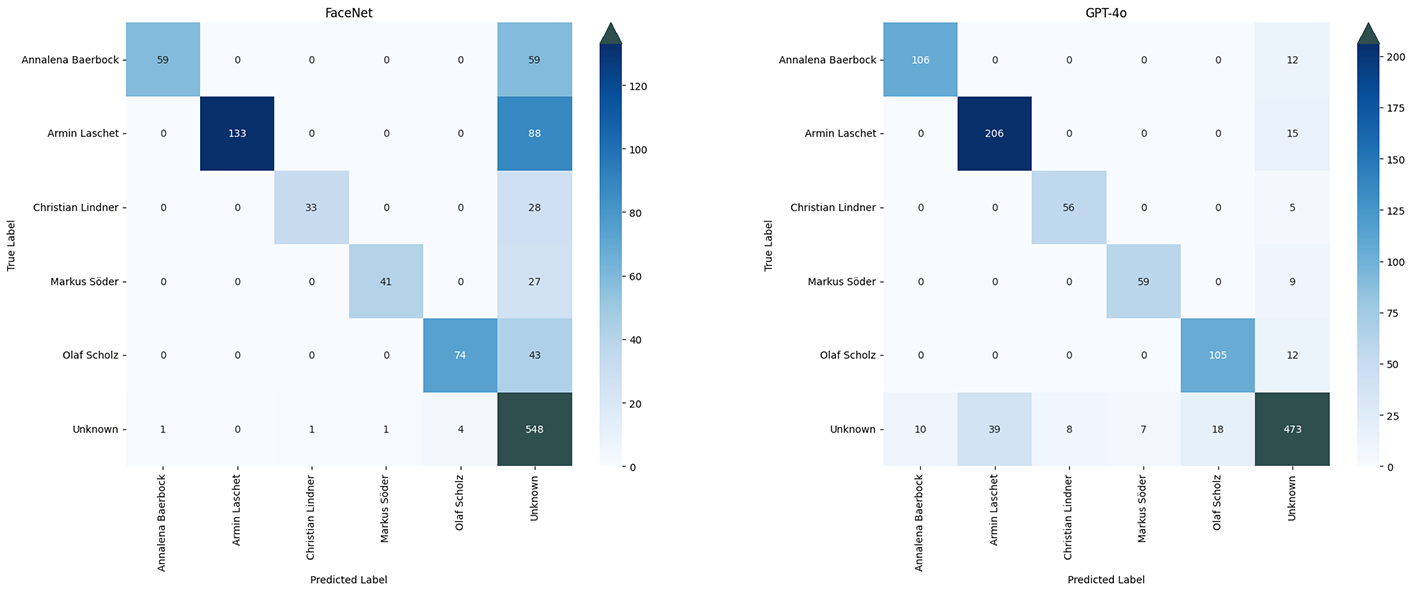}
    \caption{Confusion matrices comparing the computational classifications (predicted labels) with the human annotations (true labels) for stories. The left matrix displays the deep learning FaceNet model result, and the right the GPT-4o results.}
    \label{fig:verification-confusion-matrix}
\end{figure*}

The confusion matrices in Figure~\ref{fig:verification-confusion-matrix} illustrate the differences in classification performance between the two models for stories. \texttt{GPT-4o} shows fewer misclassifications and better alignment with human annotations compared to \texttt{FaceNet512}.

Krippendorff’s $\alpha$ was calculated to measure interrater agreement, incorporating each model as an additional annotator. \texttt{GPT-4o} achieved higher agreement with human annotations ($\alpha = 0.85$ for stories and $0.90$ for posts) than \texttt{FaceNet512} ($\alpha = 0.62$ for stories and $0.84$ for posts), indicating that \texttt{GPT-4o} aligns more closely with human-level consistency.

\begin{figure}
    \centering
    \includegraphics[width=1\linewidth]{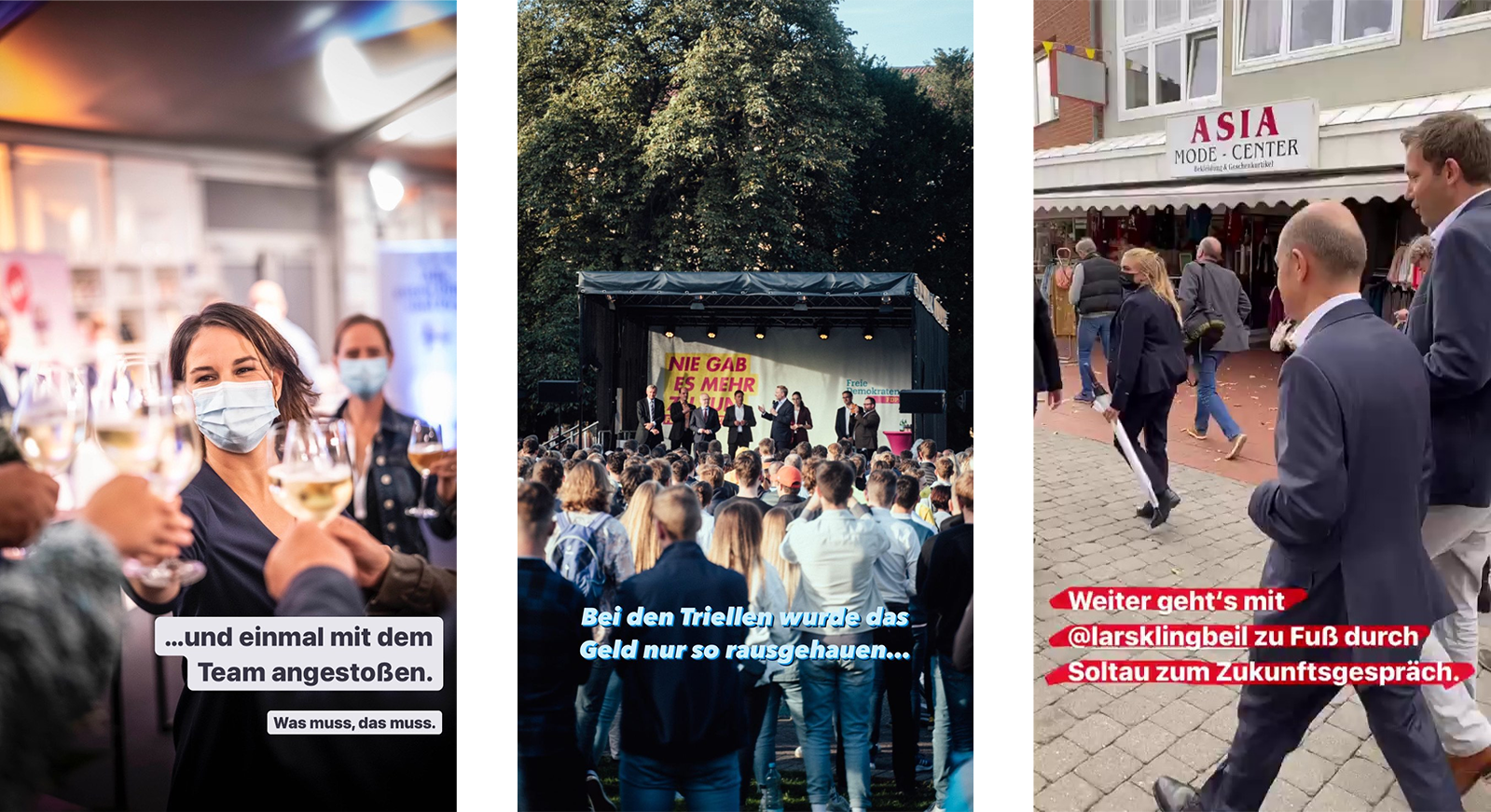}
    \caption{Several examples of stories where the model missed recognizing the faces: On the left Annalena Baerbock wearing a face-mask, in the centre Christian Lindner from the distance, yet visible to the human eye, on the right Olaf Scholz from the side.}
    \label{fig:face-problems}
\end{figure}

A qualitative exploration highlights some problems both models face: images where candidates wore face masks, were photographed from a distance, or appeared in motion led to misclassification (see Figure~\ref{fig:face-problems} for examples). While these observations are identical across both models, there is one peculiarity with images and GPT-4o: In a few instances, the model did not identify the front-runners in high-quality portrait images. The comment generated by the model usually stated a variation of ``I cannot identify individuals in images.'' Thus, the alignment of the model prevents the identification.

\subsection{RQ1b: Computational Person Count}
To address RQ1b, we compared the performance of three computational models—\texttt{DeepFace}, \texttt{GPT-4o}, and \texttt{GCV}—in counting the number of visible people in Instagram stories and posts. Using human annotations as ground truth, we evaluated each model’s precision, recall, and $F_1$-scores to determine which provides more reliable results. For each image, the model predicted the number of people visible, categorized into four classes: 0, 1, 2, or 3+ individuals. Correct counts contributed to both recall and precision, while overcounts (false positives) and undercounts (false negatives) negatively affected these metrics.

Table~\ref{tab:face-count} summarizes the overall evaluation metrics for all models across stories and posts. The \texttt{GPT-4o} model outperformed both \texttt{DeepFace} and \texttt{GCV}, achieving higher macro and weighted $F_1$-scores, especially in stories. Specifically, \texttt{GPT-4o} attained a macro $F_1$-score of 0.86 for stories and 0.93 for posts, compared to \texttt{DeepFace}’s 0.48 and 0.53, and \texttt{GCV}’s 0.58 and 0.44, respectively.

\begin{table}[ht]
\centering
\caption{Comparison of the evaluation results for the person count using different models. The table is split into \textbf{P}osts and 
\label{tab:face-count}
\textbf{S}tories. We calculated Krippendorff's $\alpha$ to measure interrater agreement, incorporating the model as a fourth annotator to compare our results with social science research. }
\centering
\resizebox{\columnwidth}{!}{%
\begin{tabular}{l | c | c | c | c || c | c |}
\toprule
\textbf{Model} & \multicolumn{2}{c |}{\textbf{Macro $F_1$}} & \multicolumn{2}{c ||}{\textbf{Weighted $F_1$}}  & \multicolumn{2}{c |}{\textbf{Kripp. $\alpha$}} \\
      & \textbf{S}         & \textbf{P}  & \textbf{S}         & \textbf{P}  &   \textbf{S}         & \textbf{P}       \\
\midrule
GPT-4o & \textbf{0.86}  & \textbf{0.93} &  \textbf{0.93}  & \textbf{0.92} & \textbf{0.82} & 0.85   \\    
GCV & 0.58 & 0.44 &  0.58  &  0.50 & 0.59 & 0.57 \\
RetinaFace      & 0.48 & 0.53 & 0.63   &  0.73 & 0.63 & 0.73 \\
\midrule
\multicolumn{5}{r ||}{\textbf{Human Annotations}} &  0.81    &     \textbf{0.86}    \\
\bottomrule
\end{tabular}
}%
\end{table}

Examining the performance per class (see Table~\ref{tab:face-count-detail} for stories), we observed that \texttt{GPT-4o} consistently provided higher precision and recall across all person count categories. For instance, for images containing precisely two people, \texttt{GPT-4o} achieved an $F_1$-score of 0.96, significantly higher than \texttt{DeepFace}’s 0.54 and \texttt{GCV}’s 0.41. This improvement is notable given that both deep learning models struggled to accurately count the number of people when more than one individual was present.

\begin{table*}[ht]
\centering
\caption{Comparison of \texttt{DeepFace}, \texttt{GPT-4o}, and \texttt{GCV} models verifying front-runners in stories. \textbf{Bold} values indicate the best-performing metric for each category and metric type.}
\label{tab:face-count-detail}
\resizebox{\textwidth}{!}{%

\begin{tabular}{ l | c c c || c c c || c c c | c |}
\toprule
\multirow{2}{*}{\textbf{Person \#}} & \multicolumn{3}{c||}{\textbf{DeepFace}} & \multicolumn{3}{c||}{\textbf{GPT-4o}} & \multicolumn{3}{c|}{\textbf{GCV}} & \multirow{2}{*}{\textbf{Support}} \\
                & \textbf{Precision} & \textbf{Recall} & \textbf{F1-Score} & \textbf{Precision} & \textbf{Recall} & \textbf{F1-Score} & \textbf{Precision} & \textbf{Recall} & \textbf{F1-Score} &  \\
\midrule
0               & 0.00 & 0.00 & 0.00  & \textbf{1.00} & 0.43 & 0.60  & 0.80 & \textbf{0.57} & \textbf{0.67}  & 7 \\
1               & 0.83 & 0.63 & 0.72  & 0.93 & \textbf{0.98} & \textbf{0.95}  & \textbf{0.94} & 0.50 & 0.65  & 90 \\
2               & 0.57 & 0.51 & 0.54  & \textbf{0.96} & \textbf{0.96} & \textbf{0.96}  & 0.47 & 0.36 & 0.41  & 47 \\
3+              & 0.52 & 0.88 & 0.66  & \textbf{0.94} & 0.92 & \textbf{0.93}  & 0.44 & \textbf{0.94} & 0.60  & 49 \\
\midrule
\textbf{Macro Avg}    & 0.48 & 0.51 & 0.48  & \textbf{0.96} & \textbf{0.82} & \textbf{0.86}  & 0.66 & 0.59 & 0.58  & 193 \\
\textbf{Weighted Avg} & 0.66 & 0.64 & 0.63  & \textbf{0.94} & \textbf{0.94} & \textbf{0.93}  & 0.69 & 0.58 & 0.58  & 193 \\
\bottomrule
\end{tabular}
}%
\end{table*}

\begin{figure*}
    \centering
    \includegraphics[width=1\linewidth]{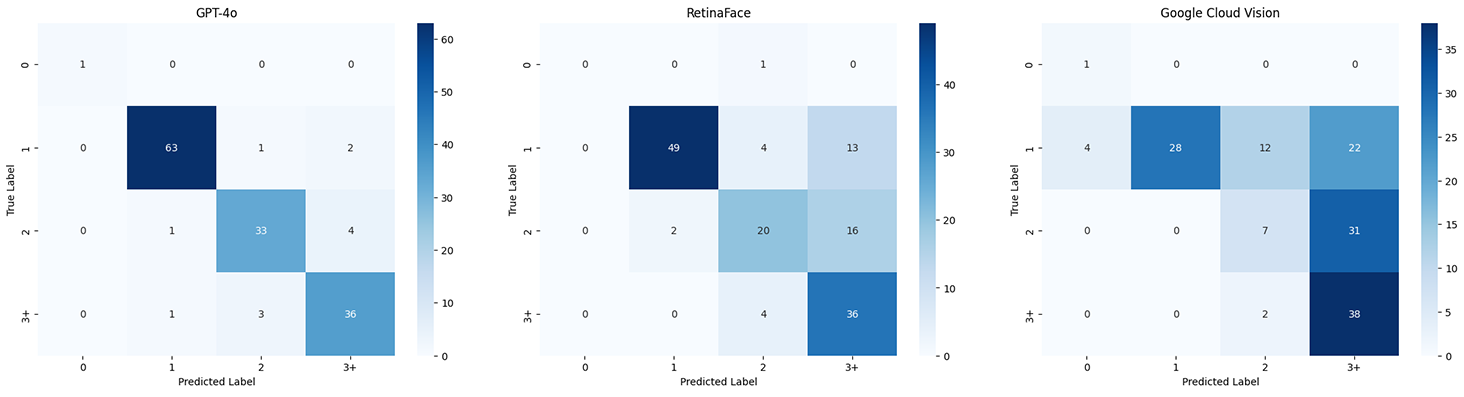}
    \caption{Three confusion matrices comparing the person count performance across the four labels and between the three classification approaches, here for posts. The deep learning model appears to overcount.}
    \label{fig:confusion-count-posts}
\end{figure*}

The confusion matrices in Figure~\ref{fig:confusion-count-posts} illustrate the differences in counting performance between the three models for posts. Both \texttt{DeepFace} and \texttt{GCV} tend to overcount the number of people, often misclassifying images with fewer individuals as having more. In contrast, \texttt{GPT-4o} demonstrates a more balanced performance, with fewer misclassifications and better alignment with human annotations.

Krippendorff’s $\alpha$ was calculated to measure interrater agreement, incorporating each model as an additional annotator. \texttt{GPT-4o} achieved higher agreement with human annotations ($\alpha = 0.82$ for stories and $0.85$ for posts) than \texttt{DeepFace} ($\alpha = 0.63$ for stories and $0.73$ for posts) and \texttt{GCV} ($\alpha = 0.59$ for stories and $0.57$ for posts), indicating that \texttt{GPT-4o} aligns more closely with human-level consistency.

A qualitative exploration highlights some challenges all models face: images with crowded scenes, individuals partially obscured, or people in the background led to miscounts. While these issues affected all models, \texttt{DeepFace} and \texttt{GCV} often overcounted due to detecting faces in backgrounds or on posters, whereas \texttt{GPT-4o} was better at focusing on the main subjects of the images.

\begin{figure}
    \centering
    \includegraphics[width=1\linewidth]{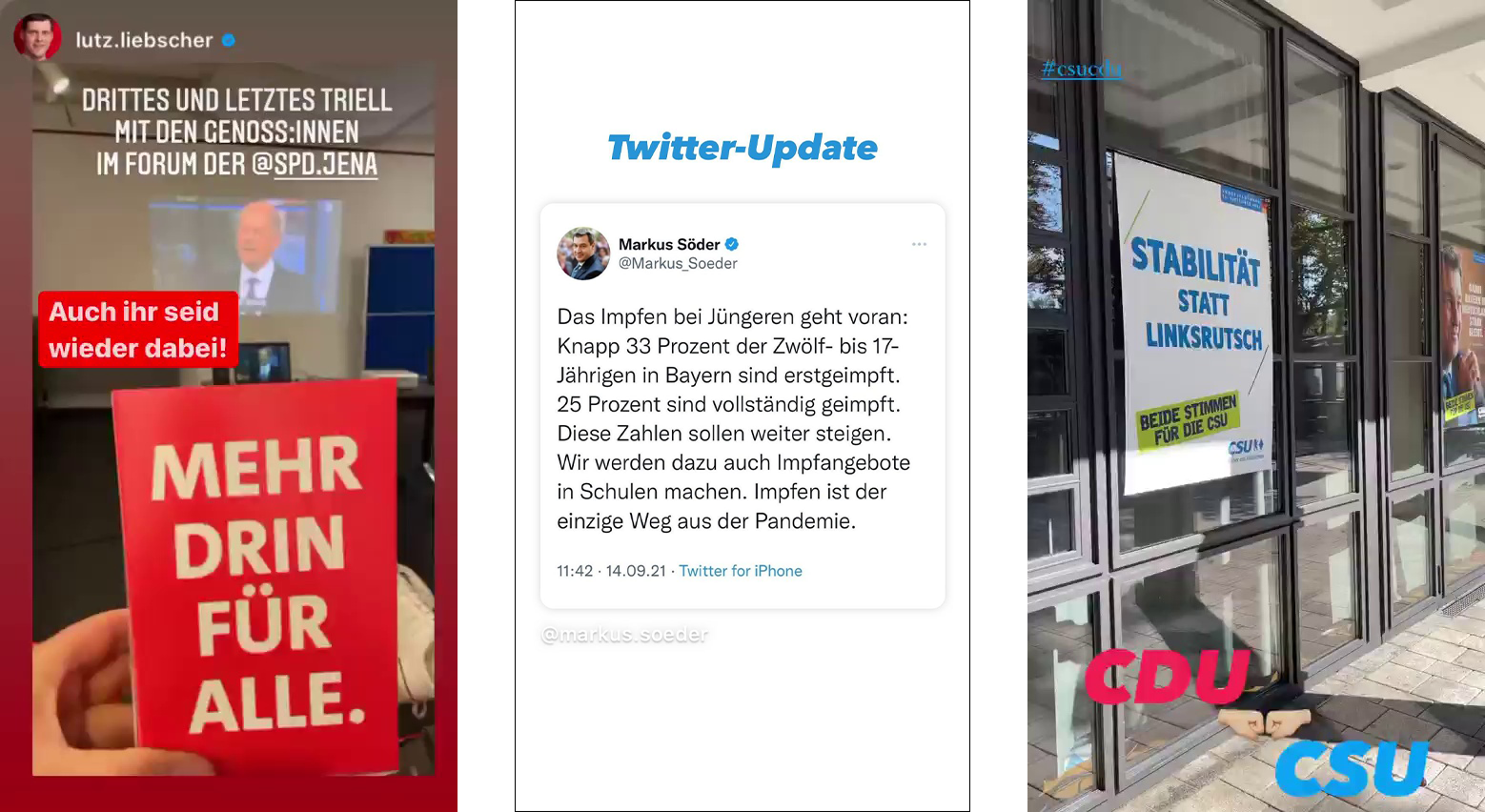}
    \caption{Three examples of images where the \texttt{RetinaFace} model detected faces. In the face recognition stage, human annotators identify the faces. When counting persons, human annotators agreed that the correct person count is 0.}
    \label{fig:zero-faces}
\end{figure}

It is important to note that only images where \texttt{RetinaFace} detected faces and the annotators identified the presence of candidates were used in the evaluation of the counting models. Despite this, some images were annotated by humans as containing zero people (compare Figure~\ref{fig:zero-faces}). These images typically included merchandise featuring the candidates or screenshots of social media posts where the candidate's profile image was visible. While \texttt{RetinaFace} detected faces in these images (e.g., on posters or screens), human annotators correctly identified them as containing zero actual people. This discrepancy introduces a bias in evaluating the zero-person class. Consequently, the macro $F_1$-scores may not fully reflect the models' performance across all classes. To mitigate this bias, we added the weighted $F_1$-scores to tables~\ref{tab:face-count} and~\ref{tab:face-count-detail}, which account for the class distribution and provide a more balanced evaluation.

Overall, our findings indicate that the multimodal \texttt{GPT-4o} model provides a more accurate and reliable method for counting people in images compared to traditional deep learning approaches. The ability of \texttt{GPT-4o} to interpret visual content holistically allows it to avoid common pitfalls such as overcounting background figures or misinterpreting pictures of faces, resulting in performance that closely aligns with human annotations.

\section{Results: Case Study}
In this section, we present the findings from our case study analysis of concentrated visibility in visual political communication. The focus is on identifying patterns of visual emphasis on front-runner candidates, examining variations across party and candidate accounts, and comparing Instagram stories and posts. Each research question is addressed in turn, starting with the visual focus on front-runners.

\subsection{RQ2a: Visual Focus on Front-Runners Across Accounts and Parties}
To address RQ2a, we analyzed how the patterns of visual focus on front-runners vary between party accounts and front-runner (candidate) accounts across different political parties on Instagram. We categorized each story and post image into three categories based on the presence of the front-runner: `No Candidate' ($C_0$), `Candidate Only' ($C_1$), and 'Candidate + Others' ($C_{+}$) This categorization allowed us to quantify the extent to which accounts focus visually on front-runners and to compare these patterns across different parties and account types.

Table~\ref{tab:overall-table} summarizes the distribution of front-runner appearances in stories and posts for both party and candidate accounts.

\begin{table}[ht] 
\centering 
\caption{Distribution of Front-Runner Appearances in Stories and Posts by Account Type} 
\label{tab:overall-table} 
\resizebox{\columnwidth}{!}{%
\begin{tabular}{l | c c c | c c c} 
\toprule 
& \multicolumn{3}{c |}{\textbf{Stories (\%)}} & \multicolumn{3}{c}{\textbf{Posts (\%)}} \\
\textbf{Account Type} & $C_0$ & $C_1$ & $C_{+}$ & $C_0$ & $C_1$ & $C_{+}$ \\
\midrule 
Party & 73.4 & 12.0 & 14.7 & 47.6 & 22.0 & 30.4 \\
Candidate & 39.2 & 33.5 & 27.2 & 22.5 & 34.0 & 43.5 \\
\bottomrule 
\end{tabular} 
}%
\end{table}

\begin{figure*}
    \centering
    \includegraphics[width=1\linewidth]{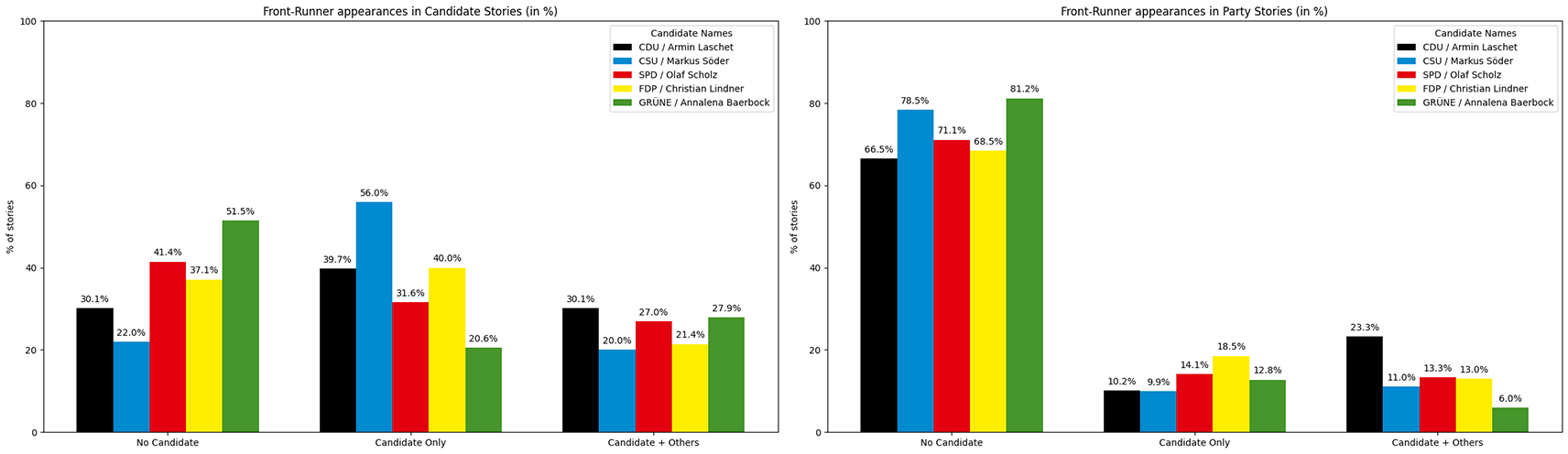}
    \caption{Comparison of front-runner appearances in Instagram \textbf{stories}, grouped into candidate and party accounts.}
    \label{fig:descriptive-stories}
\end{figure*}

\begin{figure*}
    \centering
    \includegraphics[width=1\linewidth]{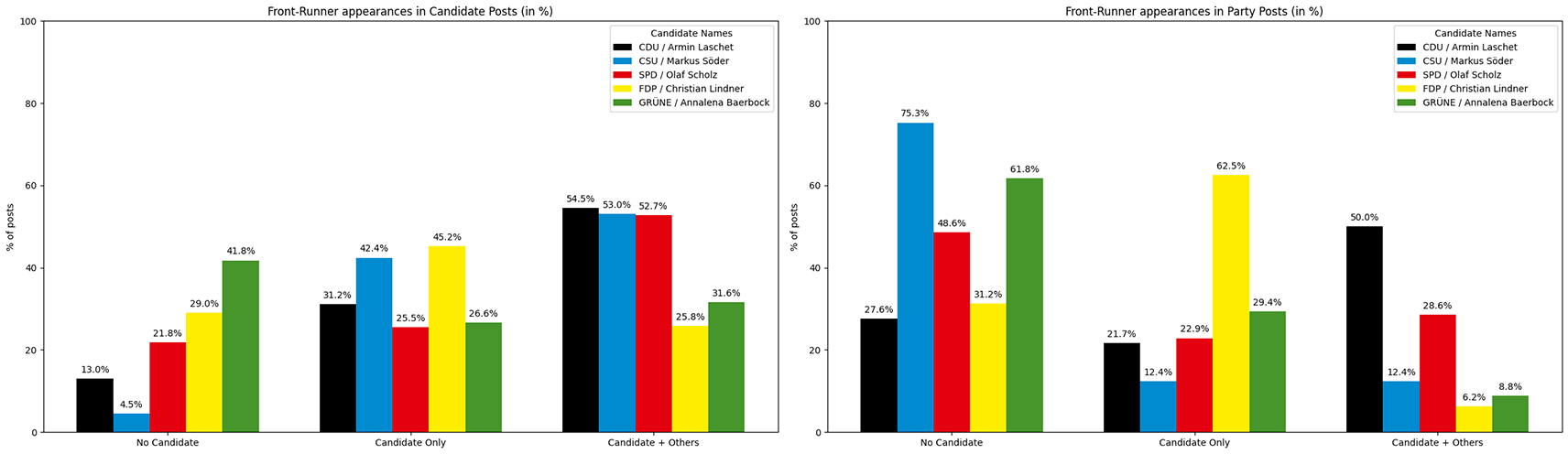}
    \caption{Comparison of front-runner appearances in Instagram \textbf{posts}, grouped into candidate and party accounts.}
    \label{fig:descriptive-posts}
\end{figure*}

A chi-squared test revealed significant differences between party and candidate accounts in the portrayal of front-runners, both in stories ($\chi^2 = 173.24$, $p < .001$, Cramér's $V = 0.349$) and posts ($\chi^2 = 48.49$, $p < .001$, Cramér's $V = 0.262$). This indicates that political communication strategies differ substantially depending on whether a party or a front-runner shares content Figure~\ref{fig:descriptive-stories} illustrates the differences in the visual focus on front-runners across parties and account types in stories, Figure~\ref{fig:descriptive-posts} in posts.


By examining individual parties, we observed that candidate accounts generally feature front runners more prominently than party accounts. For instance, Markus Söder (CSU) appears alone in 56.0\% of his stories, whereas the CSU party account shows him alone in only 13.7\% of stories. Similarly, Armin Laschet (CDU) appears alone in 39.7\% of his own stories but only in 18.0\% of the CDU's stories.

To assess the statistical significance of these differences, we conducted chi-squared tests for each party, applying Bonferroni corrections for multiple comparisons. Table~\ref{tab:chi_squared_tests} presents the results. The results show significant differences between party and candidate accounts for most parties in stories. The strongest association is observed for the CSU (Cramér's $V = 0.53$ in stories and $V = 0.697$ in posts), indicating a substantial divergence in visual strategies. The FDP shows a significant difference in stories ($p = .014$) but was excluded from the posts analysis due to insufficient data.

In posts, significant differences are found for the CSU and SPD, suggesting that these parties' candidate and party accounts employ different visual strategies even in static content.

\begin{table*}
\centering
\caption{Chi-Squared Tests for Candidate vs. Party accounts in Stories and Posts by Party. Bonferroni corrected. We have not observed enough posts by FDP to conduct a $\chi^2$ test.}
\label{tab:chi_squared_tests}
\begin{tabular}{l | ccc | ccc }
\toprule
                 & \multicolumn{3}{c |}{\textbf{Stories}}   & \multicolumn{3}{c }{\textbf{Posts}}    \\ 
\textbf{Party}   & \textbf{$\chi^2$} & \textbf{p-value} & \textbf{Cramér's V} & \textbf{$\chi^2$} & \textbf{p-value} & \textbf{Cramér's V} \\
\midrule
CDU              & 71.87             & $< .001$***                & .40                & 6.88              & .160                       & .17               \\
B90/Die Grünen   & 34.13             & $< .001$***                & .32                & 7.05              & .147                       & .25               \\
CSU              & 63.52             & $< .001$***                & .53                & 79.09              & $< .001$***                        & .70               \\
FDP              & 11.70             & .014*                      & .31                & ---            &   ---              & ---               \\
SPD              & 27.22             & $< .001$***                & .31                & 10.66             & .024*                      & .29               \\
\bottomrule
\end{tabular}
\end{table*}

\subsection{RQ2b: Visual Focus Differences Between Stories and Posts}

To address RQ2b, we examined how the patterns of visual focus on front-runners differ between Instagram stories and posts across all political parties and account types. We compared the distribution of front-runner appearances in stories and posts and conducted chi-squared tests to assess the significance of any differences observed.

Overall, a significant difference was found in the portrayal of front-runners between stories and posts ($\chi^2 = 95.20$, $p < .001$, Cramér's $V = 0.212$). This suggests that stories and posts serve different roles in the communication strategies of political parties and candidates. 

We further analyzed the differences by account type. For candidate accounts, the difference in visual focus between stories and posts was significant ($\chi^2 = 37.18$, $p < .001$, Cramér's $V = 0.191$), indicating that candidates adjust their visual strategies depending on the format. For party accounts, the difference was even more pronounced ($\chi^2 = 72.03$, $p < .001$, Cramér's $V = 0.255$), highlighting that parties may use stories and posts differently to achieve their communication objectives. Table~\ref{tab:chi_squared_stories_posts} provides the chi-squared test results for each party when comparing stories and posts.

\begin{table}
\centering
\caption{Chi-Squared Tests for Stories vs. Posts by Party. Bonferroni corrected.}
\label{tab:chi_squared_stories_posts}
\begin{tabular}{l| ccc }
\toprule
\textbf{Party}   & \textbf{$\chi^2$} & \textbf{p-value} & \textbf{Cramér's V} \\
\midrule
CDU              & 53.97             & $< .001$***                & .28               \\
B90/Die Grünen   & 8.79              & .012                        & .14               \\
CSU              & 19.50             & $< .001$***                & .22               \\
FDP              & 9.65              & .008*                      & .22               \\
SPD              & 17.56             & $< .001$***                & .21               \\
\bottomrule
\end{tabular}
\end{table}

The CDU, CSU, and SPD show significant differences in the representation of front-runners between stories and posts, with moderate effect sizes. This indicates that these parties strategically use the two formats to vary their visual focus on front-runners. In contrast, B90/Die Grünen and FDP exhibit less pronounced differences. The FDP's difference is significant but with a smaller effect size (Cramér's $V = 0.219$), suggesting a more consistent visual strategy across formats.

\section{Discussion}
This section synthesizes the findings from our study, addressing both the evaluation of computational models and the case study analysis of concentrated visibility. The first subsection focuses on the performance and implications of the evaluated models, while the second examines the insights gained from applying these methods to Instagram stories and posts during the 2021 German federal election.

\subsection{Model Evaluation}
Our comparative model evaluation results have several implications. \textbf{First}, prompting multimodal models like GPT-4o show strong potential for analyzing visual political communication: With macro averages of 0.88 for precision and 0.89 for recall, GPT-4o achieves an error rate of approximately 11.5\% in identifying front-runners in stories. Given that human annotators also showed high—but not perfect—agreement, these results highlight the challenges of achieving absolute consistency even among experts. The performance of \texttt{GPT-4o} was remarkably strong, especially compared to \texttt{FaceNet512}, positioning the model as a reliable tool for visual political analysis. Additionally, prompting is more accessible than the open-source face detection model, which requires extensive preparation and access to powerful hardware. Researchers can use interactive interfaces like ChatGPT to refine prompts:

Low-quality or blurry images of faces within the image background negatively impacted face verification performance in our experiments. Through our iterative prompt refinement, the language model appears to have picked up the notion of ``clearly depicts'' and ``clearly visible''. Implementing such a differentiation in the \texttt{deepface} workflow would have required additional work. 
In contrast, a few words in the prompt solved the problem.\footnote{We did not conduct systematic tests whether this term positively affects the classification outcome. Through our iterative prompt development, the text, including the ``clearly visible'' notions, showed the best results for a small sample.}

\textbf{Second}, caution is needed. \texttt{GPT-4o} as a closed-source model raises concerns about transparency and replicability. We cannot eliminate the possibility that our dataset has been included in the training data of \texttt{GPT-4o}, which could potentially bias the results. Future studies should conduct experiments with new datasets and open-source models to validate our results and explore the applicability of prompting across a broader range of theoretical models, such as visual framing. 

\textbf{Third}, classification quality was generally better for posts than stories, likely due to differences in content type. This difference may indicate substantial variations between the two content types. A qualitative inspection of post images suggests that they consist of high-quality, professional photos that consistently adhere to a uniform form and format. At the same time, stories contain less polished photos, including a larger share of video snapshots. Future research could use theoretical frameworks, such as image type analysis, to better understand the characteristics of images shared in posts versus stories and how these affect computational analysis.

\textbf{Fourth}, the \texttt{FaceNet} model performed worst for Annalena Baerbock, the only female candidate in our corpus. This result can be interpreted from two perspectives. 
\begin{enumerate*}[label=(\alph*)]
    \item previous research has shown that computer vision models often exhibit biases related to gender, with reduced accuracy for female faces and individuals from underrepresented groups \cite{Schwemmer2020-qn}. It is possible that we encountered such a bias here, which may have negatively affected the model's performance for the female candidate.
    \item Baerbock's style of personalization may differ from that of her male counterparts, influencing the model's performance \cite{Magin2024-cv,Geise2024-un}. She has been shown to use deviating visual frames in posts \cite{Steffan2023-pd}, which may affect face verification negatively. Additionally, qualitative observations suggest that Baerbock is often pictured wearing face masks—a notable aspect of the 2021 election during the COVID-19 pandemic.

\end{enumerate*}

Expanding the dataset to include more candidates of different genders will help future work to determine whether the observed differences are due to systematic biases in the models or to variations in how politicians visually personalize their campaigns.

\textbf{Finally}, the approach we presented using \texttt{GPT-4o} demonstrates the potential of multimodal LLMs as reliable tools for analyzing visual political communication. By enabling automated analysis with minimal technical preparation, it offers a practical framework for researchers in the social sciences. At the same time, our study highlights important trade-offs between accessibility, performance, and data sensitivity. While models like \texttt{GPT-4o} and Google Cloud Vision offer ease of use, they rely on third-party servers, raising privacy concerns. Open-source models, such as \texttt{FaceNet} and \texttt{RetinaFace}, support local processing and offer greater control over sensitive data; however, they require specialized technical expertise and suitable hardware. Future work could therefore also explore the use of open-source multimodal LLMs, such as the Llama vision models, which could combine the benefits of local data processing with the ease of prompting. However, these models currently demand powerful hardware, posing a potential barrier to broader adoption. In contrast, using GPT-4o required no dedicated infrastructure and remained relatively affordable, with a typical classification request priced at approximately \$0.01. Recognizing these trade-offs can help researchers make informed methodological decisions based on the specific needs and constraints of their projects. While \citeauthor{Bene2024-sj} (\citeyear{Bene2024-sj}) frame this argument in the context of policy issue communication, their call for more extensive comparative research across national contexts extends naturally to the visual layer of political communication as well. By lowering technical and financial barriers for large-scale visual analyses, the approach introduced here has the potential to contribute to realizing this goal.  Looking forward, it may serve as a foundation for more complex analyses of visual framing and self-presentation across both ephemeral and permanent formats.

\subsection{Case Study}
Our findings show significant differences in concentrated visibility between Instagram stories and posts, highlighting distinct strategies for each format. Consistent with previous research \cite{Hasler2021-gg}, we observed a complementary strategy between front-runners and their party accounts, particularly in stories. However, unlike \citeauthor{Hasler2021-gg}'s analysis, this complementarity is less apparent in posts. Our focus on the image level provides more detail on visual communication but complicates direct comparisons with this earlier study. Overall, candidate profiles were the primary platform for self-presentation during the 2021 election across both content types.

The underuse of concentrated visibility in party account stories suggests a missed opportunity to align with platform practices and user expectations as outlined in the theory section \cite{Parmelee2023-ft, Georgakopoulou2021-pc}. Parties could have leveraged this feature to engage voters more effectively, especially younger voters.

However, our analysis was limited to the front-runner candidate for each party, which may not fully capture the extent of personalization efforts across all candidates. Future studies could expand this scope to include a broader range of political figures to better understand the dynamics of personalization in social media campaigns. Additionally, applying theoretical frameworks like visual framing could provide deeper insights into the strategies behind image selection and presentation. 

\subsection{Limitations}
Our study faced several limitations. First, the analysis focused only on front-runner candidates, potentially limiting the generalizability of our findings. Including more candidates from different parties and genders would provide a more comprehensive picture of personalization efforts. Second, \texttt{GPT-4o}, a proprietary model, presents challenges related to transparency and replicability. Third, the dataset's limited diversity may have introduced biases, as observed by the the performance of \texttt{FaceNet} for the only female candidate. Fourth, the COVID-19 pandemic context, including the frequent use of face masks, posed additional challenges for face recognition, particularly for FaceNet. Lastly, our corpus is limited to two weeks due to the ephemeral nature and prospective data collection. 

\subsection{Future Work}
Future research should expand the scope beyond front-runner candidates to include a wider array of political figures, enabling a more comprehensive understanding of personalization dynamics. Investigating other theoretical constructs like visual framing and self-presentation could provide deeper insights into image selection and presentation strategies. Moreover, longitudinal studies could examine how these strategies evolve over time and in different electoral contexts. Further research should also explore alternative computational models and techniques to improve the transparency and replicability of findings. 

\section{Conclusion}
In conclusion, our comparative evaluation of automated methods for measuring concentrated visibility in political campaigns on Instagram highlighted the potential of prompting and multimodal large language models for analyzing visual political communication. Our findings reveal distinct strategies behind Instagram stories and posts, with candidates leveraging stories for personal branding while party accounts focus less on front-runners. Future research should address the identified limitations and further explore the use of computational tools for analyzing visual political content.

\section{Ethics}
This study collected publicly available Instagram content from official political accounts using automated scraping techniques. In accordance with European Union and German copyright exceptions (§60d UrhG; Directive (EU) 2019/790), data collection was conducted solely for non-commercial, scientific research purposes. Human annotators, recruited from student and staff communities, voluntarily participated in annotation tasks and were either reimbursed or compensated with participant hours. Annotators were assigned anonymized IDs to calculate interrater agreement; no personal data beyond publicly known figures and official account metadata was collected. Ethical standards for research integrity were maintained throughout the study.

Due to copyright, we cannot share the visual corpus publicly. Annotations, model outputs, prompts, and all evaluation notebooks are openly available at \url{https://github.com/michaelachmann/concentrated-visibility}.

\section{Acknowledgments}
We thank Philip Pirkl for exploring several approaches to improve the face verification performance in his Bachelor's thesis and for helping organize annotation studies. 
As reported, we have used ChatGPT and the GPT-API in our study. Additionally, as non-native English authors, we appreciated the support from ChatGPT and Grammarly in correcting spelling and grammar mistakes, suggesting alternative wording, and improving the overall text quality.

\bibliography{literature} 

\appendix
\section{Appendix}

\begin{table*}[ht]
\centering
\caption{Comparison of the \texttt{FaceNet512} and \texttt{GPT-4o} model verifying front-runners in posts. \textbf{Bold} values indicate the best-performing metric for each category and metric type.}
\label{tab:face-verification-detail-posts}
\begin{tabular}{ l | c c c || c c c | c |}
\toprule
\multirow{2}{*}{\textbf{Person}} & \multicolumn{3}{c||}{\textbf{FaceNet512}} & \multicolumn{3}{c|}{\textbf{GPT-4o}} & \multirow{2}{*}{\textbf{Support}} \\
                & \textbf{Precision} & \textbf{Recall} & \textbf{F1-Score} & \textbf{Precision} & \textbf{Recall} & \textbf{F1-Score} &  \\
\midrule
Annalena Baerbock     & \textbf{0.92} & 0.76 & 0.83 & 0.89 & \textbf{0.83} & \textbf{0.86} & 59 \\
Armin Laschet         & \textbf{0.99} & 0.79 & 0.88 & 0.97 & \textbf{0.98} & \textbf{0.97} & 177 \\
Christian Lindner     & \textbf{1.00} & 0.71 & 0.83 & 0.92 & \textbf{0.96} & \textbf{0.90} & 55 \\
Markus Söder          & \textbf{1.00} & 0.87 & 0.93 & 0.96 & \textbf{0.92} & \textbf{0.94} & 87 \\
Olaf Scholz           & \textbf{0.99} & 0.87 & \textbf{0.93} & 0.89 & \textbf{0.91} & 0.90 & 79 \\
Unknown               & 0.73 & \textbf{0.98} & 0.84 & \textbf{0.86} & 0.89 & \textbf{0.88} & 251 \\
\midrule
\textbf{Macro Avg}    & \textbf{0.94} & 0.83 & 0.87 & 0.92 & \textbf{0.90} & \textbf{0.91} & 708 \\
\textbf{Weighted Avg} & 0.90 & 0.87 & 0.87 & 0.91 & \textbf{0.91} & \textbf{0.91} & 708 \\
\bottomrule
\end{tabular}
\end{table*}

\begin{figure*}
    \centering
    \includegraphics[width=1\linewidth]{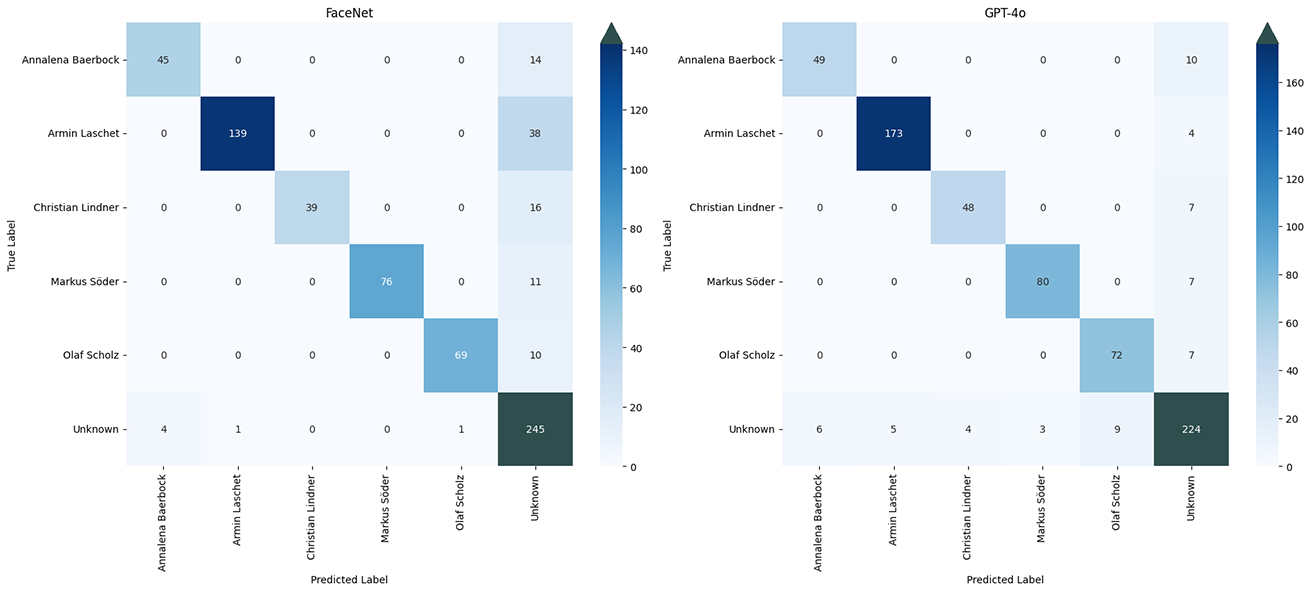}
    \caption{Confusion matrices comparing the computational classifications (predicted labels) with the human annotations (true labels) for posts. The left matrix displays the deep learning FaceNet model result, and the right the GPT-4o results.}
    \label{fig:confusion-face-verification-posts}
\end{figure*}

\begin{table*}[ht]
\centering
\caption{Comparison of \texttt{DeepFace}, \texttt{GPT}, and \texttt{GCV} for counting faces (posts). \textbf{Bold} values indicate the best-performing metric for each category and metric type.}
\label{tab:face-count-detail-posts}
\resizebox{\textwidth}{!}{%
\begin{tabular}{ l | c c c || c c c || c c c | c |}
\toprule
\multirow{2}{*}{\textbf{Person \#}} & \multicolumn{3}{c||}{\textbf{DeepFace}} & \multicolumn{3}{c||}{\textbf{GPT}} & \multicolumn{3}{c|}{\textbf{GCV}} & \multirow{2}{*}{\textbf{Support}} \\
                & \textbf{Precision} & \textbf{Recall} & \textbf{F1-Score} & \textbf{Precision} & \textbf{Recall} & \textbf{F1-Score} & \textbf{Precision} & \textbf{Recall} & \textbf{F1-Score} &  \\
\midrule
0               & 0.00 & 0.00 & 0.00  & 1.00 & 1.00 & 1.00  & 0.20 & 1.00 & 0.33  & 1 \\
1               & 0.96 & 0.74 & 0.84  & 0.97 & 0.95 & 0.96  & 1.00 & 0.42 & 0.60  & 66 \\
2               & 0.69 & 0.53 & 0.60  & 0.89 & 0.87 & 0.88  & 0.33 & 0.18 & 0.24  & 38 \\
3+              & 0.55 & 0.90 & 0.69  & 0.86 & 0.90 & 0.88  & 0.42 & 0.95 & 0.58  & 40 \\
\midrule
\textbf{Macro Avg}    & 0.55 & 0.54 & 0.53  & 0.93 & 0.93 & 0.93  & 0.49 & 0.64 & 0.44  & 145 \\
\textbf{Weighted Avg} & 0.77 & 0.72 & 0.73  & 0.92 & 0.92 & 0.92  & 0.66 & 0.51 & 0.50  & 145 \\
\bottomrule
\end{tabular}
}%
\end{table*}

\begin{figure*}
    \centering
    \includegraphics[width=1\linewidth]{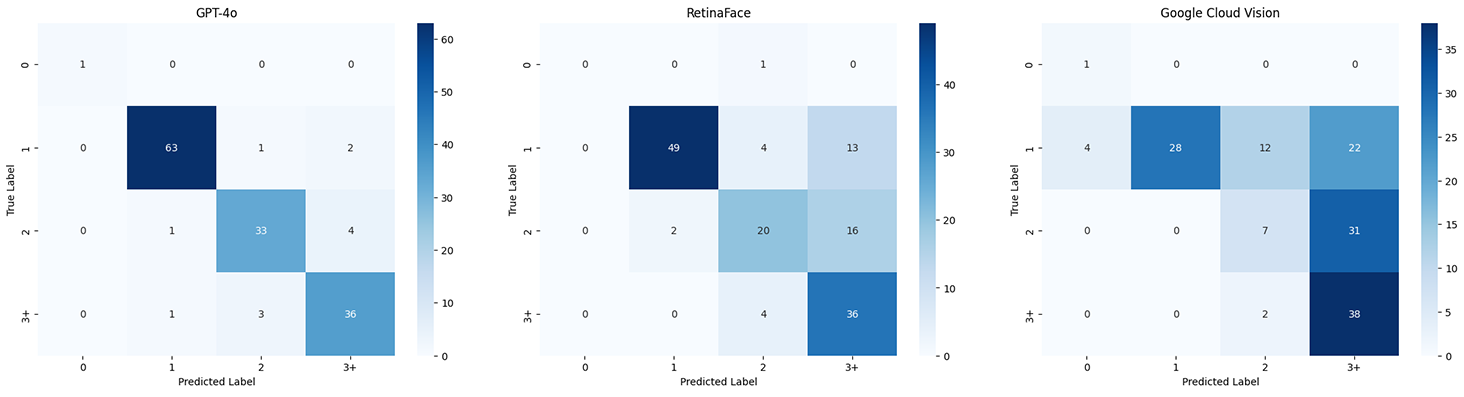}
    \caption{Three confusion matrices comparing the person count performance across the four labels and between the three classification approaches, here for stories.}
    \label{fig:confusion-count-stories}
\end{figure*}

\end{document}